# Contextual Sense Making by Fusing Scene Classification, Detections, and Events in Full Motion Video

Marc Bosch, Joseph Nassar, Benjamin Ortiz, Brendan Lammers, David Lindenbaum, John Wahl, Robert Mangum, and Margaret Smith

*Accenture Federal Services, Arlington, VA, USA.*

**Abstract**: With the proliferation of imaging sensors, the volume of multi-modal imagery/video far exceeds the ability of human analysts to adequately consume and exploit it. Full motion video (FMV) possesses the extra challenge of containing large amounts of redundant temporal data. We aim to address the needs of human analysts to consume and exploit data given aerial FMV. We have investigated and designed a system capable of detecting events and activities of interest that deviate from the baseline patterns of observation given FMV feeds. We have divided the problem into three tasks: (1) Context awareness, (2) object cataloging, and (3) salient event detection. The goal of context awareness is to constraint the problem of visual search and detection in video data. A custom image classifier categorizes the scene with one or multiple labels to identify the operating context and environment. This step helps reducing the semantic search space of downstream tasks in order to increase their accuracy. The second step is object cataloging, where an ensemble of object detectors locates and labels any known objects found in the scene (people, vehicles, boats, planes, buildings, etc.). Finally, context information and detections are sent to the event detection engine to monitor for certain behaviors and object interactions. A series of analytics monitor the scene by keeping track of object counts, and object interactions. If these object interactions are not declared to be commonly observed in the current scene, the system will report, geolocate, and log the event. Examples of events of interest include identifying a gathering of people as a meeting and/or a crowd, alerting when there are boats on a beach unloading cargo, increased count of people entering a building, people getting in and/or out of vehicles of interest, alerting when a vehicle goes from static to moving or changes speed, etc. We have successfully applied our methods on data from different sensors at different resolutions and in a variety of geographical areas. Finally, we have also transferred this technology to other platforms including low altitude unmanned aerial vehicles and wide-area motion imagery (WAMI) systems.

# I. INTRODUCTION

Overhead imagery systems capable of automatic target recognition (ATR) and/or visual sense making need to handle objects of interest with large appearance variation. Altitude, weather conditions, time of day, and camera angle among other factors. All of them have a large influence on the visual appearance of objects as they are acquired by the camera. This has a large impact on the performance of ATR algorithms. Even though during training of such algorithms, engineers try to gather representative data for many scenarios, the reality is that neural networks and similar learning-based algorithms have a hard time learning from this data all the possible visual representations of the objects. This leads to underwhelming performance of ATR algorithms under realistic conditions. In this work, we have studied this problem and address it by constraining the object appearance space that an ATR model should deal with at any given time by taking advantage of the available visual context. By context we mean all the other pixels present in the image that are not used by the model to detect an object. In other words, a neural network can learn to infer the type of scene and predict some of the factors we mentioned earlier (e.g. altitude, weather conditions, etc.) and use this information to improve its object detection performance by constraining its search space. For instance, if a model detects the aerial platform is flying at a low altitude over water, then objects like boats and other maritime vessels should be expected as opposed to vehicles, buildings, trees, and other land-based objects. In addition, the altitude information can then be used to constrain the scale of objects to expect. These constrains simplify the identification problem and can be used to resolve uncertainty or low confidence detections.

Full motion video presents the extra challenge of containing large amounts of temporal data with significant redundancy. Human analysts must (manually) summarize and annotate relevant activities or events that occur in each video, which is extremely tedious, time consuming, and typically involves multiple human analysts per video feed looking for events of interest. In general, these events of interest are known in advance, so predictive strategies can be put in place to identify such events. We have investigated and designed a system capable of detecting events and activities of interest that deviate from the baseline patterns of observation given FMV feeds in order to reduce the amount of redundant information produced by these systems as well as providing the human analyst a faster methodology to gain situational awareness. A series of analytics monitor the scene by keeping track of object counts, and object interactions. If these object interactions are not declared to be commonly observed in the current scene, the system will report, geolocate, and log the event. Examples of events of interest include identifying a gathering of people as a meeting and/or a crowd, alerting when there are boats on a beach unloading cargo, increased count of people entering a building, people getting in and/or out of vehicles of interest, alerting when a vehicle goes from static to moving or changes speed, etc.

In this work, we have designed an ATR system capable of using context to improve detection of objects and events. In addition, we have experimented with another mechanism to inject context into a model. In this case to provide descriptive information about detections and scene content rather than with the intent of constraining the detector search space.

# II. System Overview

We have divided the problem into three tasks: (1) Context awareness, (2) object cataloging, and (3) salient event detection. We have successfully applied our methods on data from different sensors at different resolutions and in a variety of geographical areas (see figure 1 for more details). Although our initial intent was to use an object detector capable to handle FMV feeds from a UAV platform in unconstrained situations, we quickly observed that due to the large complexity of the scenes, the variability in image quality and the difficulty of collecting data in realistic scenarios for, a much larger model was needed to model all possible object representations. Due to its infeasibility, a different strategy was needed to achieve satisfactory performance. We turned into a divide-and-conquer methodology where the problem and object representation could be split into semantically coherent regions. Visual context information can provide queues that enable partition of the problem space.

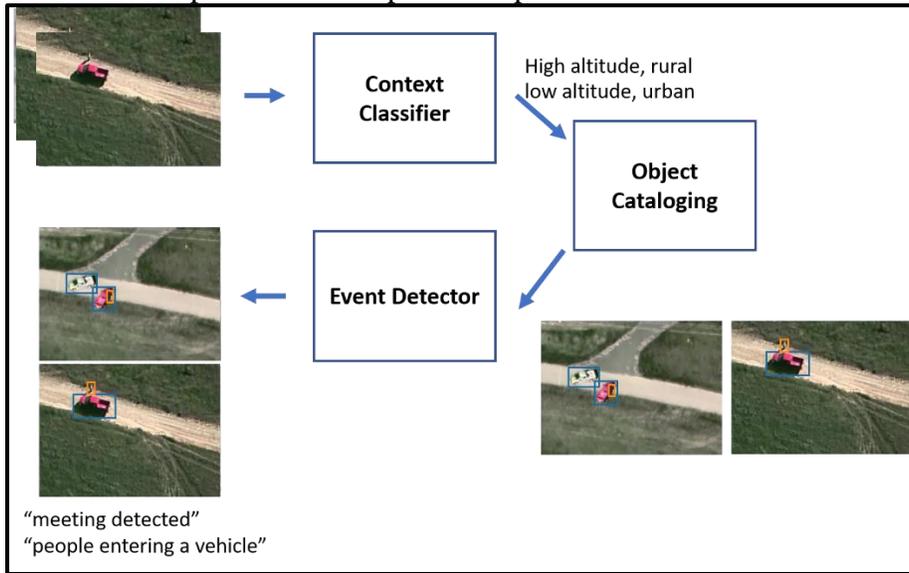

*Figure 1. Overall workflow for sense making using context.*

# III. Context Detection

Our context classifier consists of a neural network that regresses learned visual features and estimates the likelihood of each video frame to belong to a predefined scene type. Our visual context classifier, therefore, assigns a frame to sets of labels according to features extracted from the entire frame. Once the label is assigned to each frame, a set of system parameters are selected for the object cataloging module. By dynamically adjusting certain hyperparameters the object cataloging can be more precise at scanning the object representation space. In the next section we list the specific parameters tuned based on the context classifier feedback. This approach can be extended to any configurable or tunable parameter of downstream modules, including network weights, and architectures.

Without loss of generality, we used five labels for the purpose of our experiments: high altitude, medium altitude, low altitude, water, and "uneventful". Our uneventful label identified areas where the rest of the computational pipeline could be skipped due to its low likelihood for finding objects of interest given the mission.

As it can be seen in figure 2, the current design of this module consists of a ResNet50 [1] backbone feature extractor pretrained on ImageNet [7] and finetuned with our labels, and a SoftMax regressor. The regressor consisted of an average pooling layer with dropout (40%), a dense layer with SoftMax activations was used as the classifier for the 5 scene types.

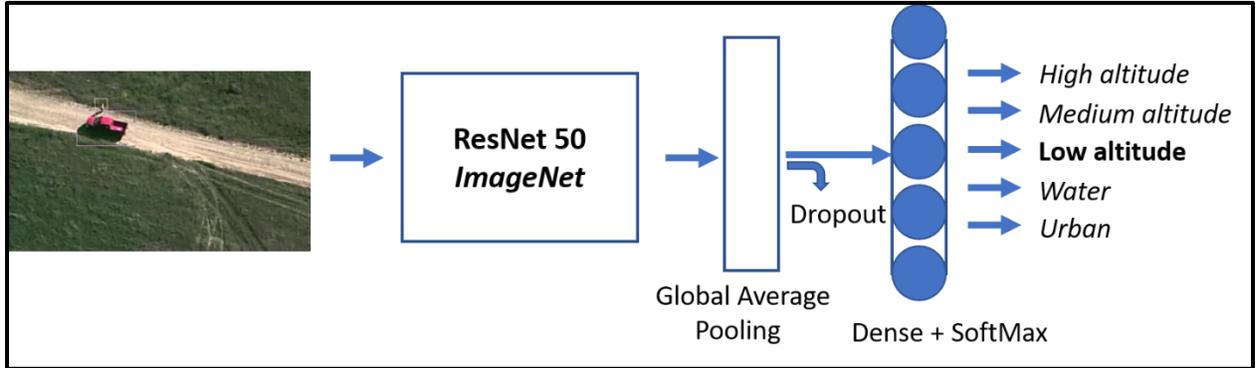

*Figure 2. Block diagram of our context classifier.*

# IV. Object Cataloging

In this section, we provide an overview of the object cataloging or detector. In order to provide a solution to the scale problem in aerial ATR scenarios, we designed a multi-scale detector where different versions of the same frame presented at different resolutions are processed by the neural network that locates the objects of interest. The collaboration between the context classifier and the multi-scale approach showed promising results. Having the information of the type of context from the previous stage, allowed us to design a model where the resolution parameter, the confidence threshold, and the minimum size of an object were tuned. The resolution parameter controls the input size of the frames being passed to the detector. The confidence threshold was used to decide on the likelihood of presence or absence of an object in each part of the frame. As an example, for scenes that no water was detected the likelihood of vessel, and boats were set to 0.9 or higher. This would force the detector to only detect boats in urban or rural environments that would exceed that threshold. As another example, the largest size in high altitude urban scenarios allowed to the detector to trigger the detection of a vehicle was much smaller than in mid or lower altitude environments. Similar reasoning was applied to other scenes and objects.

*Implementation details.* As our core component we used a RetinaNet based detector[2]. RetinaNet uses a feature pyramid network backbone on top of feedforward ResNet-50 architecture[1]. From this backbone two, subnetworks are attached: (1) box classifier – for classifying anchor boxes, (2) box regressor – for regressing from anchor boxes to ground-truth object boxes. We used this model because of relatively simplicity compared to other models at the same performance point. This one-stage detector is superior to several two-stage detectors like Faster R-CNN with feature pyramidal networks because of the loss used to train it. A focal loss function that relaxes the relative loss for well-classified examples during training by adding a factor $(1-\rho)^\gamma$ to the standard cross entropy loss. This enables to train the object detector in the presence of large number of easy background examples.
In addition to the detector, we introduced a simple object. This enabled us to build object tracks and filter the same object in consecutive frames as different detections. We implemented simple tracking-by-detection logic where detections of two consecutive frames are compared, and for each box in the current frame we examined the label id and degree of overlap with detections in the previous frame. Boxes with the same class ID and certain overlapping score - in terms of intersection over union – were merged into the same object track so that the system identify them as same object. Due to the simplicity of the approach, in crowded areas, occlusions, and for fast speed objects the method suffer considerably. One of our improvement areas is to add a track-learn-detect tracker framework for improved performance, where the tracker tracks by detection but only using the knowledge of the most current frames to learn the appearance of the object of interest.

Figure 3 shows the overall architecture used in our object cataloging.

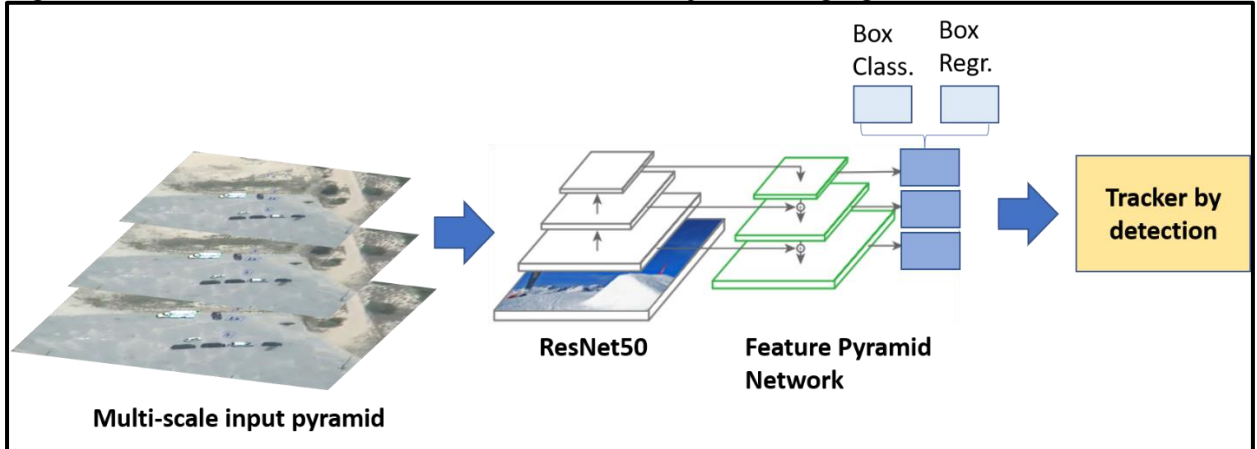

*Figure 3. Multi-scale object cataloging and tracker.*

# V. Event Detection

As we have mentioned earlier, the final component of the pipeline is an *event detector*. Its goal is to filter through the object detections identifying events of interest. These events are logged and ready for review later. These events involve activities of interest that deviate from the baseline patterns of observation given by the FMV feeds in order to reduce the amount of redundant information produced. This is also helpful for analysts since there is targeted content to inspect and review, speeding up situational awareness.

In our initial exploratory work with focused on six type of events: a person entering a vehicle, a person getting out of a vehicle, a meeting between multiple people, a crowd gathering, a person getting in a vessel, and a person getting off a vessel. We have built simple logic rules to detect these events, based on bounding box interactions. A meeting is detected when two or more people have been detected and their bounding boxes overlap or become very close in pixel count. This rule is compared with a threshold, empirically found from "baseline conditions". Similarly, crowds are reported when multiple people are detected in the same scene and their bounding boxes become proximal to their group centroid. The rest of events involved multi label logic. For instance, for a person getting in a vessel, we measure the overlap between vessel labeled boxes and people, while in the previous frame no interaction between boxes had been detected. Similarly, for someone getting out of a vehicle, if there is a large level of overlap between person and vehicle, and the person track is a new one, which means that the person is a new detection. The system resolves that as a person having got out of a vehicle.

As we add more events in our system, we are interested on having a model that can learn these object interactions from the data, rather than handcrafting detection rules. We have been focused on the development of the capability and its versatility up to this point.

# VI. Results

In this section we show some of our results, in terms of both object detection and event resolution. Figure 4 shows examples of object detections, for several types of vehicles and people. Note the difficulty of this task given the image quality and the different view points and heights that need to be accounted. Figure 5 shows examples of event detections, meeting and people exiting vehicles as a queue for start monitoring. Finally, in figure 6 we show a view of our common operating picture, where the relevant events and detections are geotagged, recorded and presented to the user for posterior analysis.

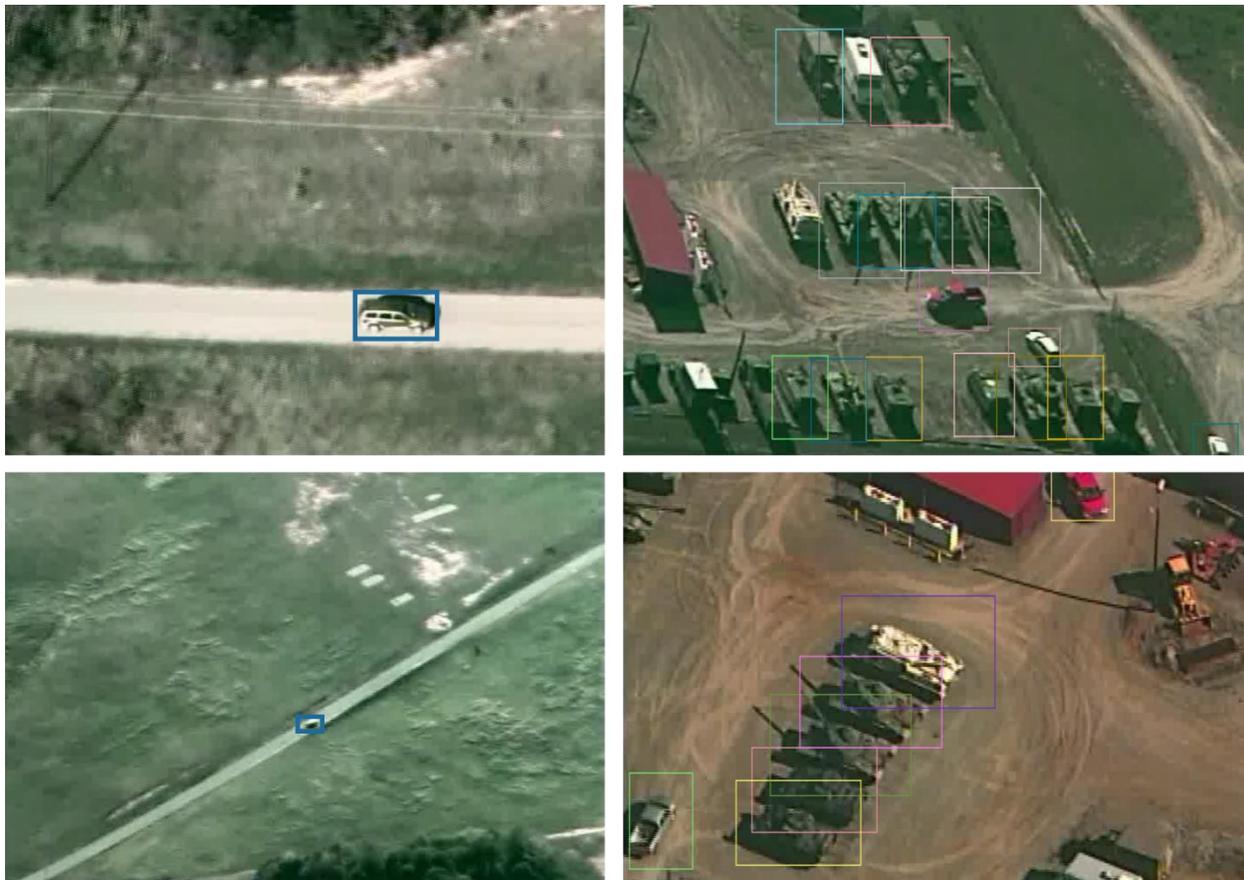

*Figure 4. Results of our context classifier and object detector. Note context classifier helps detect objects at different heights and scales (top and bottom left).*

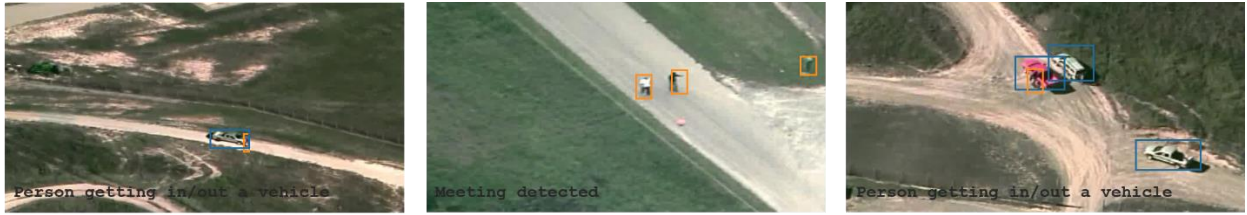
Figure 5. Event detection results. From left to right: person getting in a vehicle, meeting, person getting out of a vehicle.

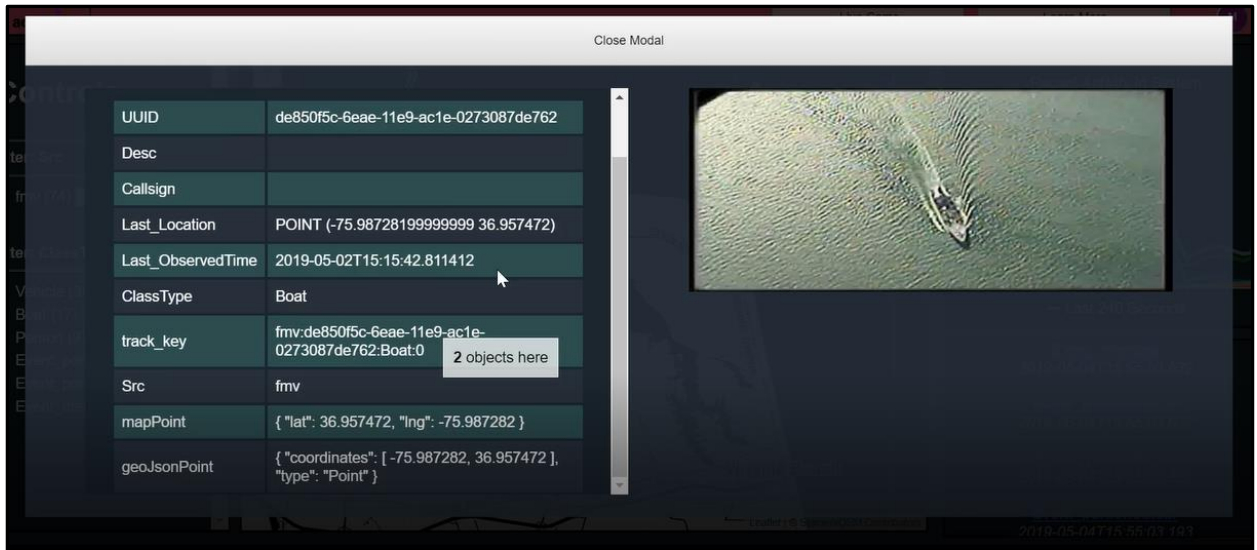
Figure 6. Common operating picture. Detection is geo-located, recorded and attributed.

# VI. Context for Object Descriptions

As mentioned earlier, we also investigated other uses of context through other available network architectures like *densecap* [3]. In this case the goal was to use context pixels to infer other information about the object. Current state-of-the-art visual perception systems for overhead imagery are limited to perceptual tasks such as object location and classification. This constrains the information provided by these systems, which are focused on solving a localization-and-labeling problem. In other words, the primary objective of these systems is to place a bounding box, or similar marker, around an object and determine the object type. However, this only answers the "where" and "what"; it does not provide information as to "why", "how", "when", or "with whom". Therefore, it misses the relationships/interactions with other objects in the scene, with object's attributes and with activities/actions. Truly capable computer vision systems need to be capable of both object detection and description. Such descriptions should capture the elements mentioned above (attributes, actions, relationships -AAR tuple). Another challenge for overhead imagery applications is that the scene is often very complex with many objects interacting with each other. For instance, current efforts on overhead image captioning are often inefficient as they can only focus on one region or describe the overall scene in very generic and limited detail, thereby limiting its value. Therefore, we believe that a complementary system to the object cataloging + event detector scheme is required. In 2016, a novel system that replaces traditional object detection (location+labeling) for a dense description of objects (location+description) was introduced [3,4]. Equally relevant was the introduction of a new dataset, *Visual Genome* [5]. A dense object description model requires a densely annotated dataset so that it can learn to reason about object labels, attributes, and relationships. *Visual Genome* enables cognitive tasks on top of perception. For each image, more than 40 descriptions of different parts of the image are annotated, textually and using a graph representation. The graph encodes relationships of the many objects in the scene, while the text descriptions enable learning models that represent such relationships and object attributes using natural language. By the numbers, *Visual Genome* contains descriptions for more than 108,000 images, with more than 255,000 objects grouped in about 18,000 object categories. Figure below shows several examples of object detectors complemented by context to produce object descriptions.

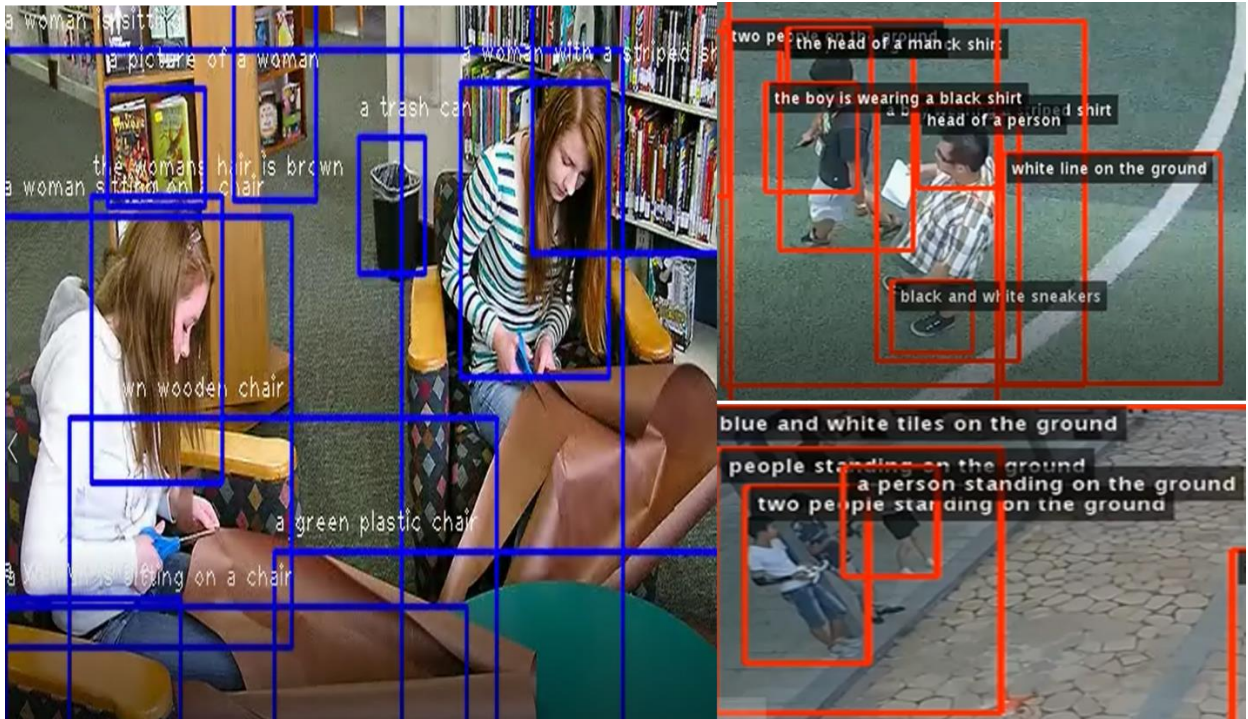

*Figure 7. Examples of dense object descriptions.*

# VII. Conclusion

In this work, we have designed an ATR system capable of using context to improve detection of objects and events. We have divided the problem into three tasks: (1) Context awareness, (2) object cataloging, and (3) salient event detection. We have described our methods for combining context pixels with state-of-the-art neural networks to constraint the problem. We can pass the context type as side information to an ATR system so it can used "optimized" parameters and settings for better performance for the given scenario/context. In addition, we have experimented with another mechanism to inject context into a model, end-to-end object description that leverages multiple subnetworks to learn object attributes, actions and relationships with other objects from context. In this case to provide descriptive information about detections and scene content rather than with the intent of constraining the detector search space.